# ChatGPT-3.5, ChatGPT-4, Google Bard, and Microsoft Bing to Improve Health Literacy and Communication in Pediatric Populations and Beyond

Kanhai S. Amin[1], Linda Mayes MD[2], Pavan Khosla BA[3], Rushabh Doshi MSc, MPH[3]
1. Yale College, New Haven, CT, USA
2. Yale Child Study Center, Yale School of Medicine, New Haven, CT, USA
3. Yale School of Medicine, New Haven, CT, USA

Corresponding Author:
Kanhai Amin, kanhai.amin@yale.edu
333 Cedar Street, New Haven, CT, 06510

There are no funding sources or conflicts of interest for this research. The authors have no declarations to make. This study did not require ethics committee review approval per 45 CFR § 46.

**Abstract:**

**Purpose**: Enhanced health literacy has been linked to better health outcomes; however, few interventions have been studied. We investigate whether large language models (LLMs) can serve as a medium to improve health literacy in children and other populations.

**Methods:** We ran 288 conditions using 26 different prompts through ChatGPT-3.5, Microsoft Bing, and Google Bard. Given constraints imposed by rate limits, we tested a subset of 150 conditions through ChatGPT-4. The primary outcome measurements were the reading grade level (RGL) and word counts of output.

**Results:** Across all models, output for basic prompts such as "Explain" and "What is (are)" were at, or exceeded, a 10th-grade RGL. When prompts were specified to explain conditions from the 1st to 12th RGL, we found that LLMs had varying abilities to tailor responses based on RGL. ChatGPT-3.5 provided responses that ranged from the 7th-grade to college freshmen RGL while ChatGPT-4 outputted responses from the 6th-grade to the college-senior RGL. Microsoft Bing provided responses from the 9th to 11th RGL while Google Bard provided responses from the 7th to 10th RGL.

**Discussion:** ChatGPT-3.5 and ChatGPT-4 did better in achieving lower-grade level outputs. Meanwhile Bard and Bing tended to consistently produce an RGL that is at the high school level regardless of prompt. Additionally, Bard's hesitancy in providing certain outputs indicates a cautious approach towards health information. LLMs demonstrate promise in enhancing health communication, but future research should verify the accuracy and effectiveness of such tools in this context.

**Implications:** LLMs face challenges in crafting outputs below a sixth-grade reading level. However, their capability to modify outputs above this threshold provides a potential mechanism to improve health literacy and communication in a pediatric population and beyond.

**Introduction**

Health literacy, which is emphasized by the American Academy of Pediatrics, Centers for Disease Control and Prevention (CDC), and Joint Commission, is a crucial component in the provision of high-quality healthcare.[1] Health literacy is defined as "the degree to which individuals have the capacity to obtain, process, and understand basic health information and services needed to make appropriate health decisions."[2] Health literacy assessment in children remains an emerging field of study; but, current estimates suggest a significant portion – up to 85% – of children and adolescents exhibit inadequate health literacy.[3,4] Enhancing health literacy in children is crucial, given its profound impact on health-related decisions, behaviors, and ensuing outcomes.

In adults, greater health literacy is linked to lower hospital admissions,[5,6] improved health status,[7] and greater understanding of chronic illnesses and their management.[8] Growing research indicates that improved health literacy has similar benefits for children and adolescents.[1] Particularly, improved health literacy in children and adolescents with chronic conditions, which affect 8-25% of children, may have significant impacts on child health during childhood and long-term, as self-care responsibilities are often transferred to the child between the ages of 11 and 15.[9,10] Notably, enhancements in health literacy have been shown to foster improved patient-provider communication, self-management, and facilitate a smoother transition to adult care for children with chronic kidney disease,[11] congenital heart disease,[12] spina bifida,[13] rheumatic conditions,[14] and cancer[15] among others.

Currently, there is limited literature analyzing the efficacy of health literacy instruments and interventions for adolescents.[16,17] Investigations of digital health interventions have gained momentum, as 75% of adolescents and young adults used the internet, primarily Google, as their most recent source of health information.[18,19] Additionally, many adolescents obtain health information from parents and educators, who frequently derive their own health knowledge from internet sources.[17]

New publicly available large language models (LLMs) such as OpenAI's ChatGPT, Google Bard, and Microsoft Bing may provide opportunities to improve health literacy in a variety of fields.[20] The implications for children are boundless, especially given that adolescents use the internet on a daily basis more than any other age group.[21,22] In this study, we assessed the ability of LLMs to explain diseases at an appropriate level when (1) a basic prompt is used and (2) when a prompt with greater context is used.

**Methods**

*Conditions:*

A comprehensive list of 288 childhood disorders and conditions encompassing a wide range of pediatric diseases including genetic abnormalities, hepatobiliary conditions, congenital irregularities, mental health issues, cardiovascular disorders, oncological cases, digestive system disorders, and dermatological conditions was compiled. The list was curated by including conditions listed on the Johns Hopkins Children's Center's[23] and Seattle Children's Hospital's[24] websites.

*Prompt Selection:*

Due to the countless number of prompts, two simple prompts "Explain {medical condition}" and "What is (/are) {medical condition}" were initially chosen, as they are expected queries of the lay individual. Two additional prompt architectures were then chosen based on the importance of context[25]: "Explain {medical condition} to a __ grader" and "Explain {medical condition} at a __-grade reading level." In these additional prompts, the grade levels first to twelfth were tested for all conditions.

*Outputs:*

We ran the 288 conditions through Open AI's ChatGPT-3.5 (5.24.23 version), Google Bard (5.23.23 version), and Microsoft Bing (5.4.23 version) for all 26 prompts. Due to current rate limits in OpenAIs ChatGPT-4 (5.23.23 version), a random sub-selection of 150 conditions was chosen to test the prompts in ChatGPT-4.

*Processing Outputs:*

To standardize and ensure equal comparison, we removed all formatting including bullet points and numbered lists, as is consistent with other studies.[26,27] Further, to compare the true output, all routine ancillary information such as "I hope this helps! Let me know if you have any other questions." and "Sure. I can help with that" were removed. Outputs unable to be generated due to LLM limitations for a particular prompt and LLM combination were excluded after one retry.

*Readability Assessment:*

We assessed the grade level of the output by using Gunning Fog (GF), Flesch-Kincaid Grade Level (FK), Automated Readability Index, and Coleman-Liau (CL) indices. Each index outputs a score corresponding to a reading grade level (RGL) i.e., a RGL of seven corresponds to the seventh-grade reading level. Along with prior literature, we averaged the four indices to find the average RGL (aRGL) of the output.[25,26] Word counts for each output were also calculated. We applied the non-parametric Wilcoxon signed-rank and rank-sum tests to compare aRGLs as appropriate. Python version 3.11 (2022) was used to gather readability scores and analysis was conducted using R (R Core Team, 2022) and RStudio (Rstudio Team, 2022)

**Results**

For the two basic prompts – "What is {}" and "Explain {}" – the aRGL was found to be at or above the high school level for all LLMs (Table 1, Fig. 1). Both ChatGPT-3.5 and ChatGPT-4 at baseline produced output at the college level (Table 1, Figure 1). Meanwhile, Bing and Bard produced output around the eleventh-grade level and the tenth-grade level, respectively (Table 1, Fig. 1). Both basic prompts performed at similar aRGLs for all LLMs besides for ChatGPT, where "explain" resulted in significantly higher aRGL output for ChatGPT-4 (p<0.0001) (Fig. 1). When comparing each LLM for the same basic prompts, differences were significant (p<0.0001) – except between ChatGPT-3.5 and ChatGPT-4 for "what is" / "what are" – with Bing and Bard at lower aRGLs than the OpenAI models (Table 1, Fig. 1). Word count varied between LLM and within LLM for basic prompts and higher aRLG did not necessarily correlate to higher or lower word count (Table 1, Fig. A1).

**Figure 1: Reading Grade Levels for Basic Prompts**
Legend: *Basic Prompts P0 "What is (are) {medical condition}" and P1 "Explain {medical condition}" were tested through the LLMs. The aRGL of outputs are shown. \*, \*\*, \*\*\*, \*\*\*\* correspond to p<0.05, p<0.01, p<0.001, and p<0.0001, respectively. Comparisons between LLM for identical prompts are not shown, but all differences are statistically significant p<0.0001, except between ChatGPT-3.5 and ChatGPT-4 for "what is" / "what are."*

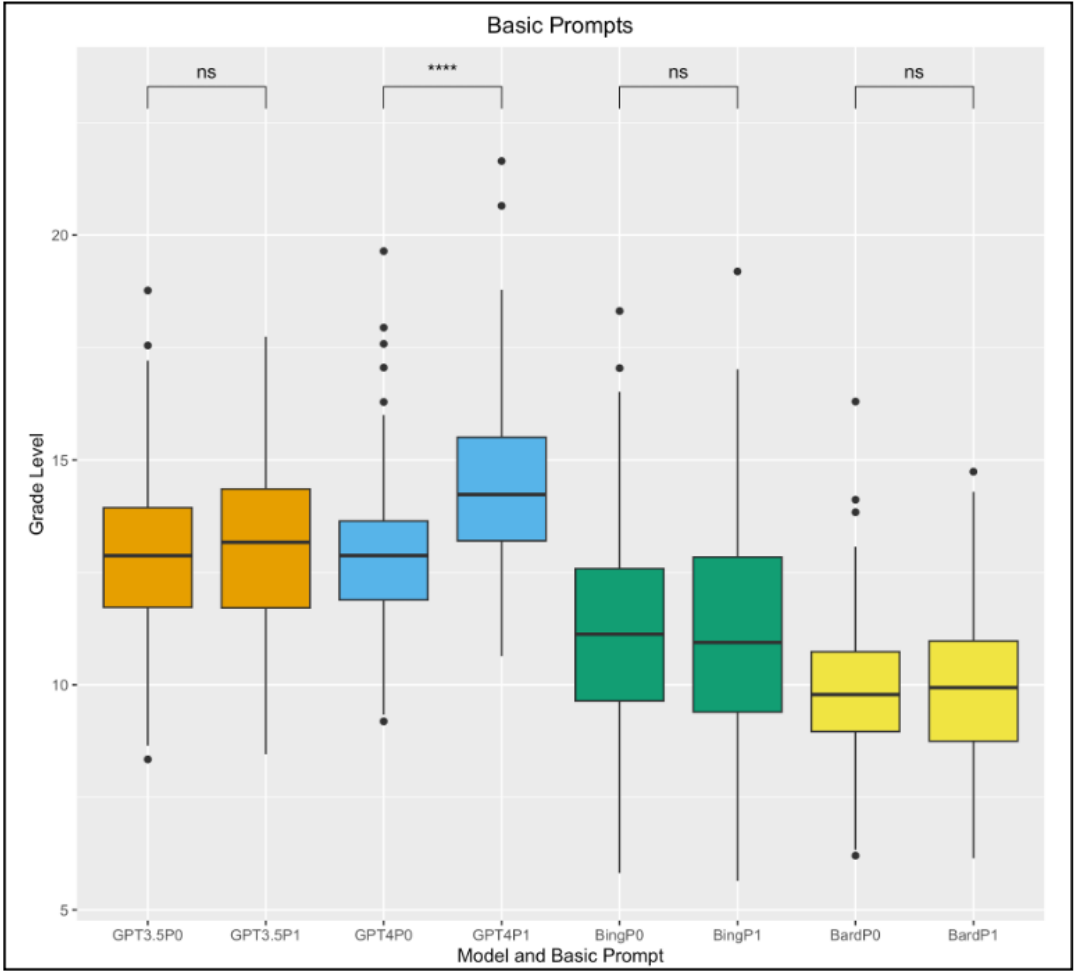

**Table 1: Reading Grade Level and Word Count for Each Prompt and LLM Combination**

Legend: Median (Quartile 1 to Quartile 3) are shown for every combination of prompt and LLM

| Prompt | ChatGPT-3.5 aRGL | ChatGPT-3.5 Word Count | ChatGPT-4 aRGL | ChatGPT-4 Word Count | Bing aRGL | Bing Word Count | Bard aRGL | Bard Word Count |
|---|---|---|---|---|---|---|---|---|
| What is / What are { } | 13.1 (11.7-14.2) | 93 (82-105) | 12.9 (11.9-13.6) | 293 (274-321) | 11.3 (9.7-12.5) | 149 (112-196) | 9.9 (9.0-11.0) | 304 (292-370) |
| Explain { } | 13.0 (12.0-14.1) | 161 (133-188) | 14.2 (13.2-15.5) | 264 (225-307) | 11.1 (9.6-12.8) | 151 (112-210) | 9.9 (8.8-11.0) | 234 (229-296) |
| Explain { } to a first grader | 7.3 (6.2-8.4) | 67 (58-83) | 7.9 (6.8-8.9) | 74 (63-87) | 9.6 (8.0-11.3) | 115 (86-157) | 7.3 (6.5-8.2) | 260 (225-289) |
| Explain { } to a second grader | 7.4 (6.5-8.5) | 75 (64-90) | 8.1 (7.3-9.5) | 85 (70-99) | 9.5 (8.0-11.3) | 114 (87-159) | 7.6 (6.7-8.4) | 256 (231-288) |
| Explain { } to a third grader | 7.8 (6.8-8.7) | 83 (71-100) | 8.7 (7.5-9.5) | 87 (74-104) | 9.7 (8.0-11.3) | 120 (90-171) | 7.8 (6.8-8.6) | 257 (245-305) |
| Explain { } to a fourth grader | 8.1 (7.2-9.1) | 93 (78-120) | 9.2 (8.2-9.9) | 98 (84-124) | 9.8 (8.2-11.4) | 124 (91-160) | 8.0 (7.1-9.1) | 277 (245-305) |
| Explain { } to a fifth grader | 8.5 (7.4-9.4) | 87 (75-102) | 9.2 (8.2-10.1) | 106 (91-138) | 10.2 (8.2-11.8) | 129 (95-182) | 8.2 (7.4-9.1) | 277 (245-311) |
| Explain { } to a sixth grader | 8.8 (7.6-9.7) | 90 (76-109) | 9.8 (8.7-10.7) | 131 (107-179) | 10.0 (8.5-11.7) | 131 (91-183) | 8.3 (7.5-9.3) | 278 (248-313) |
| Explain { } to a seventh grader | 9.3 (8.1-10.5) | 93 (80-110) | 10.5 (9.6-11.5) | 156 (120-197) | 10.3 (8.7-12.2) | 130 (95-181) | 8.6 (7.8-9.6) | 279 (254-313) |
| Explain { } to a eighth grader | 9.7 (8.7-10.9) | 91 (79-107) | 10.9 (10.2-11.7) | 158 (115-198) | 10.3 (8.8-11.9) | 134 (99-186) | 8.8 (8.0-9.8) | 289 (263-320) |
| Explain { } to a ninth grader | 10.8 (9.5-11.9) | 98 (85-124) | 11.7 (10.9-12.8) | 195 (155-233) | 10.3 (8.8-12.2) | 133 (95-180) | 9.0 (8.0-10.0) | 291 (263-324) |
| Explain { } to a tenth grader | 11.7 (10.7-12.9) | 111 (94-150) | 12.2 (11.1-13.0) | 215 (170-251) | 10.4 (8.8-12.2) | 136 (96-183) | 9.1 (8.2-10.2) | 299 (270-332) |
| Explain { } to a eleventh grader | 12.1 (10.9-13.1) | 130 (103-163) | 12.7 (12.0-13.8) | 235 (205-286) | 10.6 (9.0-12.4) | 143 (107-191) | 9.4 (8.4-10.2) | 309 (279-351) |
| Explain { } to a twelfth grader | 12.6 (11.5-13.5) | 142 (109-172) | 13.6 (12.8-14.3) | 260 (220-300) | 11.0 (9.3-12.7) | 148 (107-189) | 9.6 (8.5-10.6) | 311 (283-351) |
| Explain { } at a first-grade reading level | 7.1 (6.1-7.9) | 71 (59-82) | 6.1 (5.3-7.6) | 53 (45-67) | 9.9 (8.2-11.3) | 115 (79-183) | 7.5 (6.6-8.5) | 218 (196-246) |
| Explain { } at a second-grade reading level | 6.9 (6.0-8.1) | 74 (64-86) | 6.7 (5.5-7.5) | 66 (54-80) | 9.4 (7.9-11.1) | 126 (87-174) | 7.5 (6.6-8.7) | 233 (205-256) |
| Explain { } at a third-grade reading level | 7.5 (6.5-8.4) | 84 (69-99) | 7.3 (6.3-8.2) | 79 (67-95) | 10.1 (8.6-12.2) | 112 (82-163) | 7.9 (6.8-8.8) | 236 (210-269) |
| Explain { } at a fourth-grade reading level | 7.9 (6.9-8.8) | 92 (79-115) | 7.7 (6.7-8.6) | 96 (81-127) | 10.0 (8.4-11.7) | 108 (79-152) | 8.1 (7.0-9.1) | 249 (224-275) |
| Explain { } at a fifth-grade reading level | 8.5 (7.5-9.5) | 95 (84-118) | 8.1 (7.4-9.0) | 120 (94-161) | 10.2 (8.5-12.3) | 110 (79-162) | 8.2 (7.2-9.3) | 256 (230-286) |
| Explain { } at a sixth-grade reading level | 9.0 (7.9-10.0) | 104 (88-139) | 8.9 (8.0-9.7) | 162 (119-209) | 10.3 (8.6-12.1) | 113 (79-153) | 8.6 (7.6-9.7) | 266 (235-289) |
| Explain { } at a seventh-grade reading level | 9.6 (8.4-10.6) | 107 (87-139) | 9.8 (8.8-10.8) | 221 (176-252) | 10.5 (8.7-12.5) | 120 (89-176) | 8.8 (7.8-9.9) | 269 (242-295) |
| Explain { } at a eighth-grade reading level | 9.6 (8.7-10.7) | 106 (91-138) | 10.4 (9.7-11.3) | 239 (244-316) | 10.4 (8.9-12.1) | 122 (87-177) | 8.9 (8.0-9.8) | 267 (240-295) |
| Explain { } at a ninth-grade reading level | 10.3 (9.3-11.5) | 115 (94-147) | 11.6 (10.7-12.7) | 274 (256-320) | 10.7 (9.1-12.9) | 116 (83-168) | 9.3 (8.2-10.3) | 284 (258-321) |
| Explain { } at a tenth-grade reading level | 11.1 (10.0-12.3) | 116 (94-150) | 12.1 (11.4-13.2) | 288 (299-376) | 10.5 (9.0-12.3) | 144 (100-203) | 9.5 (8.5-10.7) | 287 (259-324) |
| Explain { } at a eleventh-grade reading level | 11.6 (10.7-12.6) | 161 (117-196) | 14.2 (12.9-15.0) | 341 (320-391) | 11.1 (9.1-12.9) | 136 (99-194) | 9.7 (8.8-10.7) | 304 (272-343) |
| Explain { } at a twelfth-grade reading level | 12.1 (11.0-12.9) | 159 (117-197) | 16.0 (14.9-16.9) | 355 (112-196) | 11.1 (9.6-12.9) | 145 (100-199) | 9.8 (9.0-10.8) | 209 (278-341) |

When adding the context of "Explain {} to a __ grader" and asked from grade 1 to 12, all LLMs struggled to reach the desired grade level output (Table 1, Fig. 2). ChatGPT-3.5 demonstrated the ability to vary median output between the seventh-grade and college freshmen aRGL, while increasing word count for higher grade-level prompts. ChatGPT-4 varied the median output between the eighth-grade level and the college sophomore reading level, while increasing word count for higher grade level prompts (Table 1, Fig. 2, Fig. A2). Microsoft Bing outputted between the tenth- and eleventh-grade aRGL, and Google Bard outputted between the seventh and tenth aRGL. ChatGPT-3.5 and ChatGPT-4 demonstrated greater variation in word count than both Bing and Bard (Table 1, Fig. A2).

*Figure 2: Reading grade level of output after running "Explain {} to a _____ grader" through each LLM*
Legend: Each LLM was asked, "Explain {medical condition} to a __ grader." First through twelfth grade were tested.
A) The aRGL of outputs is depicted for each LLM. From top to bottom, GPT-3.5, GPT-4, Bing, and Bard are depicted.
B). Grade-level outputs for each LLM from panel A are set side to side for comparison between LLMs.

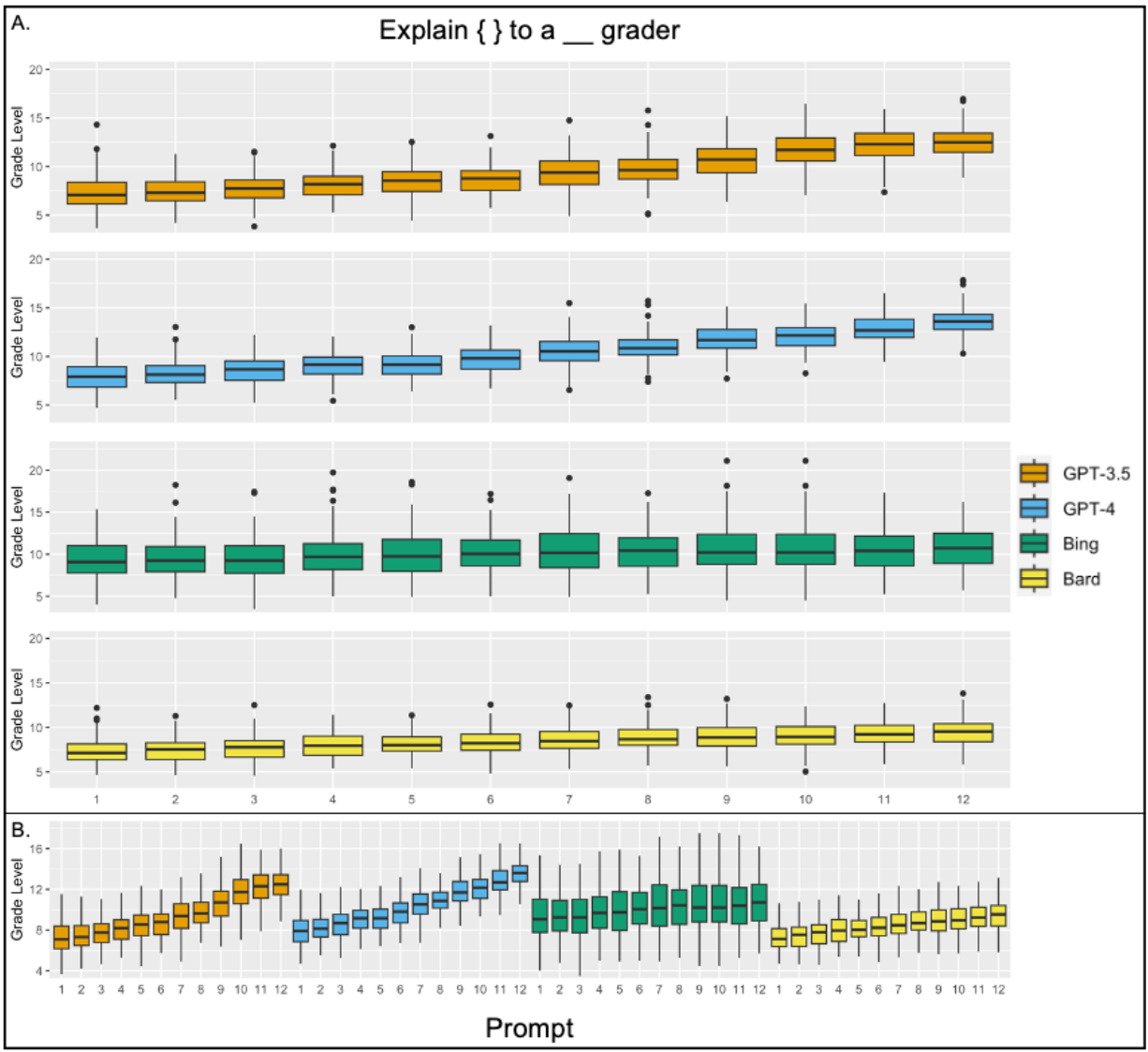

When the context was changed to a specific reading level, "Explain {} at a __-grade reading level" (assessing each condition from grade 1-12), ChatGPT-3.5 ranged output between the seventh- and twelfth-grade reading level. ChatGPT-4 varied output between the sixth-grade and college-senior reading level. Similar to the prior prompt, Bing varied output between the ninth- and eleventh-grade reading levels and Bard varied output between the seventh- and tenth-grade reading levels (Table 1, Fig. 3, Fig. A3). ChatGPT-3.5, ChatGPT-4, and Bard demonstrated increasing word count with higher grade-level prompts (Fig. A3).

*Figure 3: Reading grade level of output after running "Explain {} at a ____- grade reading level" through each LLM*

Legend: Each LLM was asked "Explain {medical condition} at a __-grade reading level." First through twelfth grade were tested. A) The aRGL of outputs is depicted for each LLM. From top to bottom, GPT-3.5, GPT-4, Bing, and Bard are depicted. B). Grade-level outputs for each LLM from panel A are set side to side for comparison between LLMs.

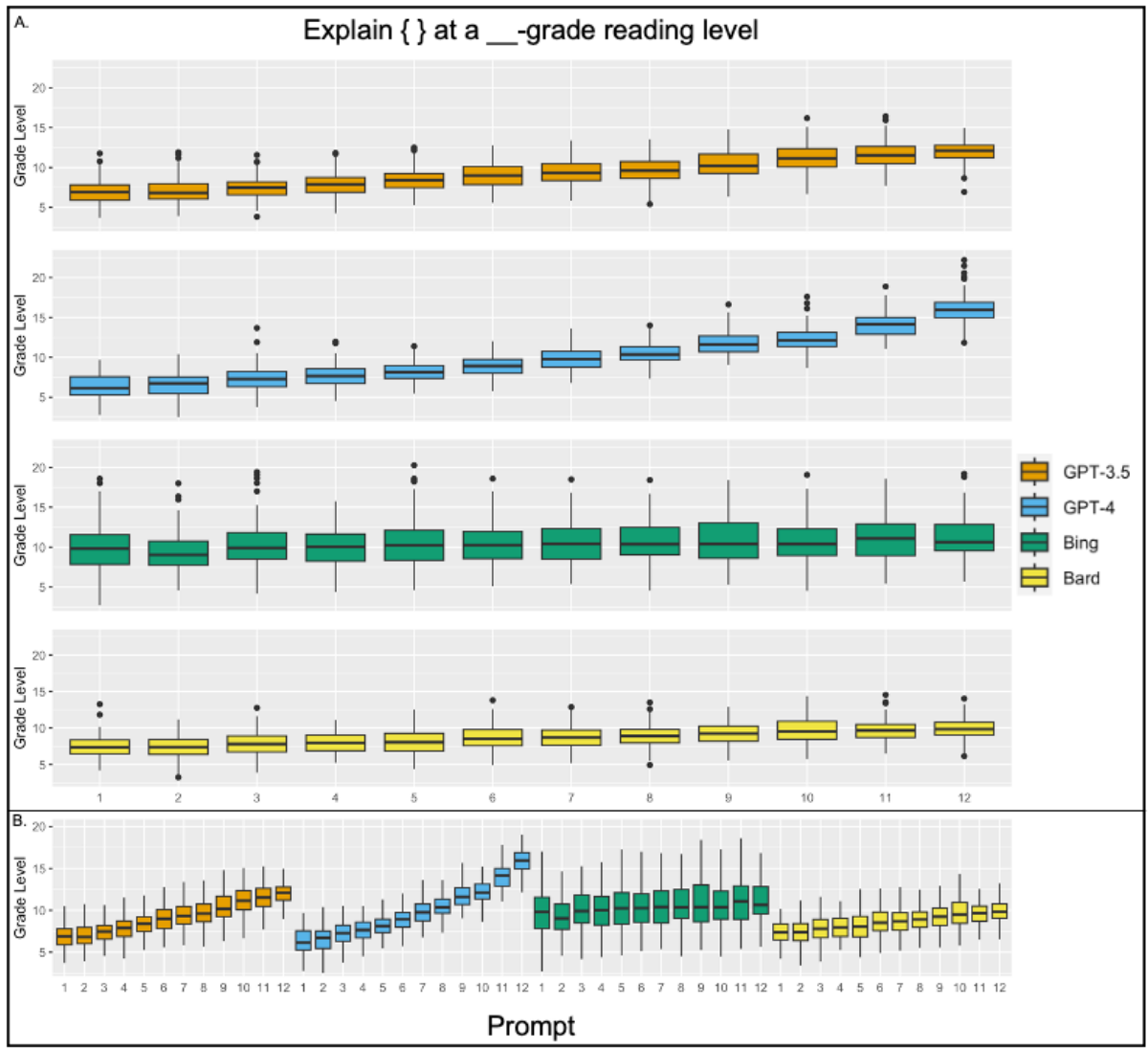

All LLMs produced output for each condition tested except Google Bard. Google Bard particularly struggled with prompts "What is" / "What are" and "Explain," failing to answer 11 and 19 conditions with the first query, respectively. For example, for these two prompts, Google Bard failed to produce output for depression and monkeypox, saying it was a limitation of being a language model, and both pectus carinatum and pectus excavatum citing language limitations.

**Discussion**

In this study, we demonstrate the current abilities and limitations of LLMs in explaining common pediatric medical conditions. Out of the numerous prompts we could have tested, we focused on two basic prompts and two prompt architectures directed towards attaining a desired reading grade level (RGL).

While no model could accurately pinpoint a desired RGL, there was a notable uptick in RGLs in parallel with an increase in prompt grade-level specification. From our study, in this specific context, OpenAI's models ChatGPT-3.5 and ChatGPT-4 demonstrated the greatest range and ability to tailor RGLs based on requested grade level, with Bard and Bing showcasing more limited range of grade levels. While OpenAI's models, ChatGPT-3.5 and ChatGPT-4, did better in achieving lower-grade level outputs, Bard and Bing tended to consistently produce an RGL that is at the high school level.

The current inability to pinpoint output to an exact RGL or generate output below the sixth-grade reading level demonstrates present limitations of LLMs. However, it is critical to note that readability scales are primarily grammatical and do not factor in context, antecedent knowledge, motivation, and informational requirements. Additionally, this limitation may exist due to training data, as most health information is at or above a high school reading level.[28,29] The training data in combination with preprocessing techniques and fundamental differences in LLM algorithms may explain the differences in performance between the LLMs.[30]

Given the importance of health literacy in children and adolescents, LLMs present a novel method to improve literacy. By adjusting the RGL, it is possible to make medical information more accessible, and therefore more comprehensible, to a wider range of readers. However, it's essential to understand that while these models can adapt their outputs to different reading levels, they are not infallible and may occasionally produce content that is either too complex or too simplistic. Hence, while LLMs like ChatGPT can be powerful tools for enhancing health literacy, they should ideally be used in conjunction with other educational tools and methods, especially when targeting pediatric populations. The practical application of LLMs could be in the creation of patient education materials that cater to various reading abilities or in generating quick explanations on medical topics that can be easily understood by children and their caregivers.

Interestingly, Bard's failure to output at initial query for certain diseases such as depression and monkeypox may represent Bard's more cautious approach towards health information.[31,32] This might reflect the developers' intent to avoid potential misinformation, particularly in a domain as sensitive as health. This cautious stance,

10while commendable, does emphasize the need for further fine-tuning to ensure relevant information isn't withheld unnecessarily. Ensuring accuracy and relevance while mitigating the risk of misinformation remains a critical challenge for LLM deployment in the healthcare sector.

Further, the interactive nature of LLMs allows patients to readily seek clarification or simplification, enhancing utility. While subsequent studies might evaluate comprehension of LLM outputs by adolescents or their parents, our findings illustrate the potential of LLMs to facilitate learning above the 6$^{th}$ grade level. As the LLMs continue to rapidly improve, their functionality as a resource for aiding parent-child communication may improve. While accuracy of the outputs were anticipated, subsequent studies should validate the accuracy, completeness, and functionality of LLMs within this context.

**Conclusion**

Adolescents and parents are increasingly expected to interact with LLMs, including ChatGPT, Bard, and Bing. As the technology continues to become more mainstream, LLMs may become a source for health information. Although the models to exhibit an effort to modulate reading grade level (RGL) in outputs, the incapacity to precisely target desired RGLs, particularly beneath a sixth grade reading level, underscores the limitations of such models. Future research is warranted to corroborate the efficacy, accuracy, and impact of LLMs in real-world healthcare communication and decision-making scenarios.

Supplemental Figures:

## AFigure1: Word Count for Basic Prompts
Legend: Basic Prompts P0 "What is (are) {medical condition}" and P1 "Explain {medical condition} were tested through the LLMs. The word count of outputs is shown.

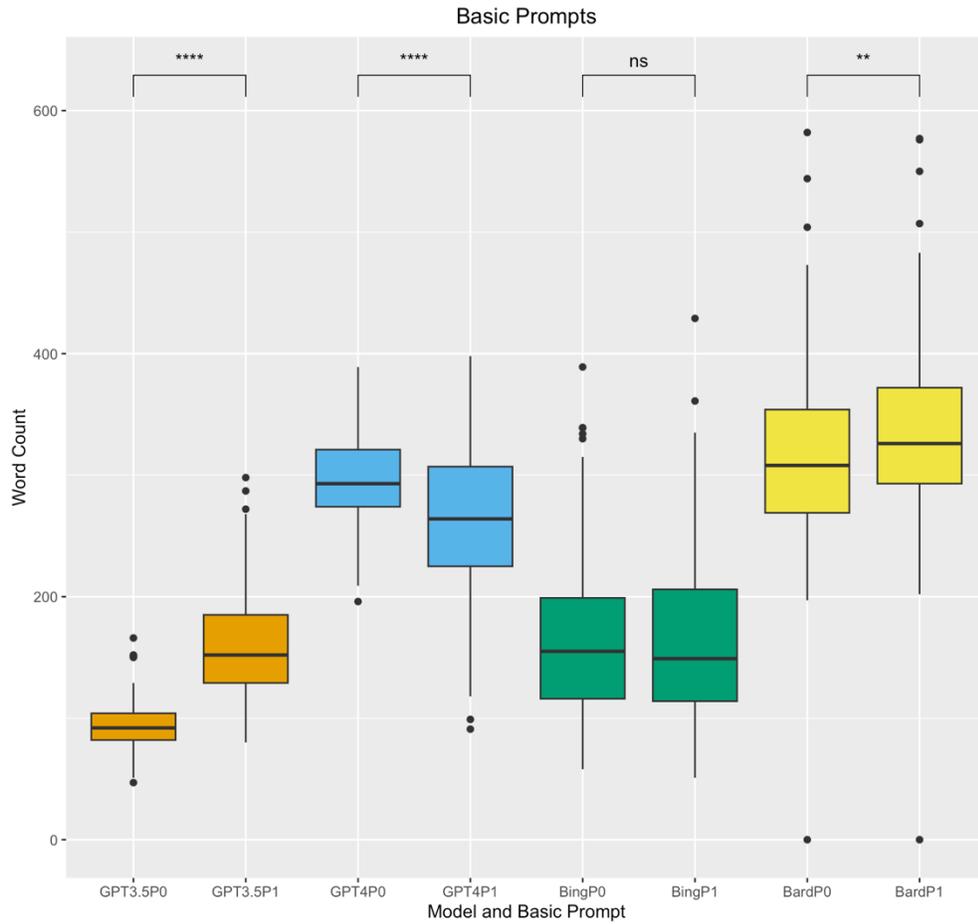

## AFigure 2: Word Count of output after running "Explain {} to a _____ grader" through each LLM

Legend: Each LLM was asked, "Explain {medical condition} to a __ grader." First through twelfth grade were tested. A) The word count of outputs is depicted for each LLM. From top to bottom, GPT-3.5, GPT-4, Bing, and Bard are depicted.  B). Word counts for each LLM from panel A are set side to side for comparison between LLMs.

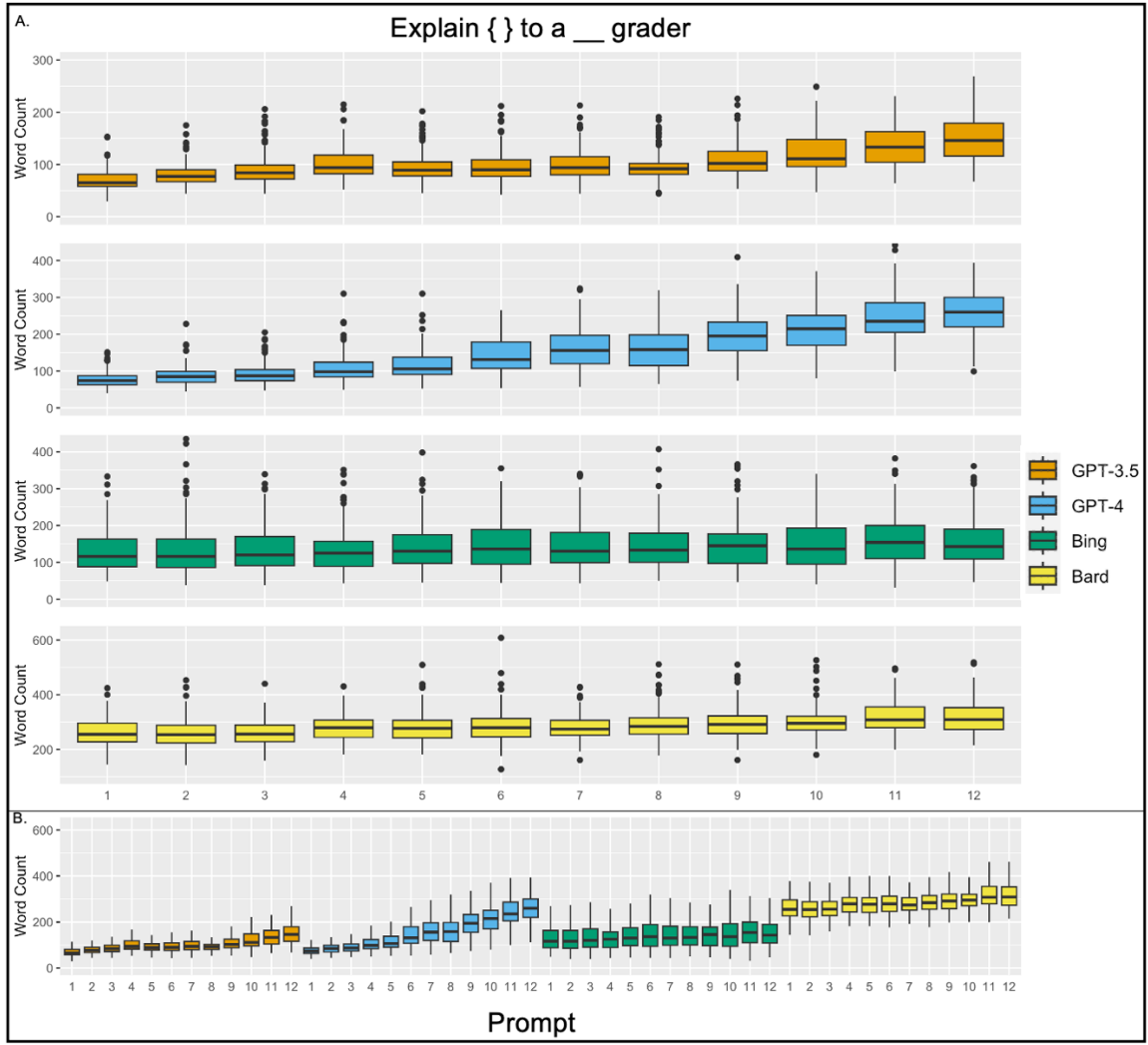

**AFigure3: Word Count of output after running "Explain {} at a ____- grade reading level" through each LLM**

Figure Legend: Each LLM was asked, "Explain {medical condition} at a __-grade reading level." First through twelfth grade were tested. A) The word counts of outputs is depicted for each LLM. From top to bottom, GPT-3.5, GPT-4, Bing, and Bard are depicted. B). Word counts for each LLM from panel A are set side to side for comparison between LLMs.

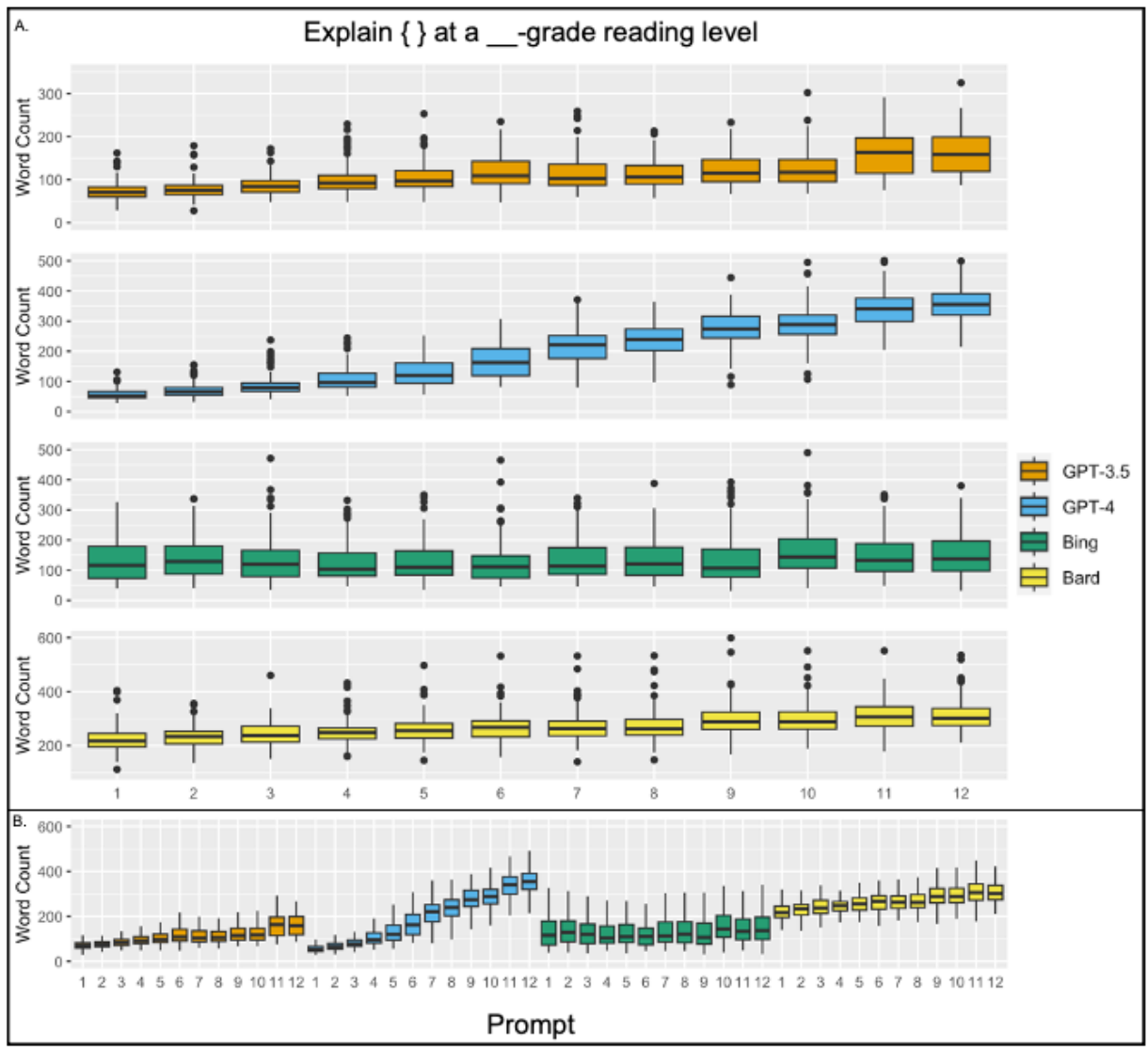